\documentclass[10pt,twocolumn,letterpaper]{article}
\pdfoutput=1

\usepackage{iccv}
\usepackage{times}
\usepackage{epsfig}
\usepackage{graphicx}
\usepackage{amsmath}
\usepackage{amstext}
\usepackage{amssymb}
\usepackage{bm}
\usepackage{multirow}
\usepackage{multicol}
\usepackage{booktabs}
\usepackage{pifont}
\usepackage{xcolor}
\usepackage{bbding} 
\usepackage{enumerate}
\usepackage{float}
\usepackage{subcaption}
\usepackage{tabularx}
\usepackage{authblk}

\usepackage[breaklinks=true,bookmarks=false]{hyperref}

\iccvfinalcopy 

\def\iccvPaperID{****} 

\ificcvfinal\fi

\begin{document}

\title{Learning Optical Flow from Event Camera with Rendered Dataset}

\author{
Xinglong Luo\textsuperscript{\rm 1,\rm 3},
Kunming Luo\textsuperscript{\rm 2},
Ao Luo\textsuperscript{\rm 3},
Zhengning Wang\textsuperscript{\rm 1},
Ping Tan\textsuperscript{\rm 2},
and Shuaicheng Liu\textsuperscript{\rm 1,\rm 3}\thanks{Corresponding author}\\
\textsuperscript{\rm 1}University of Electronic Science and Technology of China\\
\textsuperscript{\rm 2}The Hong Kong University of Science and Technology \qquad
\textsuperscript{\rm 3}Megvii Technology\\
\tt\small \{luoboom@std.,zhengning.wang@,liushuaicheng@\}uestc.edu.cn \\
\tt\small kluoad@connect.ust.hk luoao02@megvii.com pingtan@sfu.ca
}

\maketitle
\ificcvfinal\thispagestyle{empty}\fi

\begin{abstract}
We study the problem of estimating optical flow from event cameras. One important issue is how to build a high-quality event-flow dataset with accurate event values and flow labels. Previous datasets are created by either capturing real scenes by event cameras or synthesizing from images with pasted foreground objects. The former case can produce real event values but with calculated flow labels, which are sparse and inaccurate. The later case can generate dense flow labels but the interpolated events are prone to errors. In this work, we propose to render a physically correct event-flow dataset using computer graphics models. In particular, we first create indoor and outdoor 3D scenes by Blender with rich scene content variations. Second, diverse camera motions are included for the virtual capturing, producing images and accurate flow labels. Third, we render high-framerate videos between images for accurate events. The rendered dataset can adjust the density of events, based on which we further introduce an adaptive density module (ADM). Experiments show that our proposed dataset can facilitate event-flow learning, whereas previous approaches when trained on our dataset can improve their performances constantly by a relatively large margin. In addition, event-flow pipelines when equipped with our ADM can further improve performances.

\end{abstract}

\section{Introduction}
Event cameras~\cite{lichtsteiner2008128} record brightness changes at a varying framerate~\cite{brandli2014240}. When a change is detected in a pixel, the camera returns an event in the form $e=(x,y,t,p)$ immediately, where $x,y$ stands for the spatial location, $t$ refers to the timestamp in microseconds, and $p$ is the polarity of the change, indicating a pixel become brighter or darker. 
\begin{figure}[!t]
    \centering
    \includegraphics[width=\linewidth]{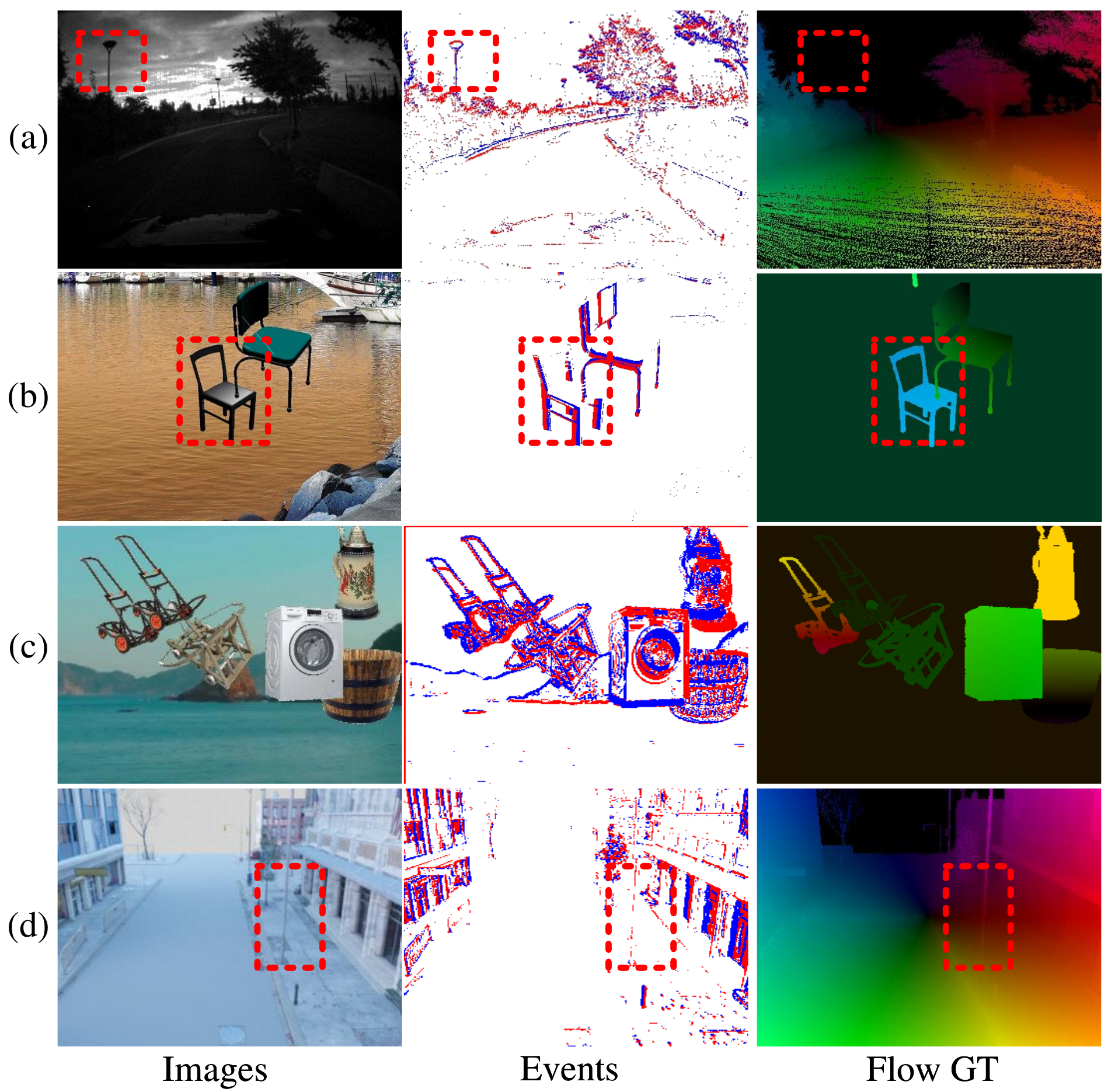}
    \caption{(a) the captured dataset from real event camera~\cite{zhu2018multivehicle,zhu2018ev}. (b) the synthesized dataset with flying chairs foreground~\cite{wan2022DCEI}. (c) the synthesized dataset by moving a foreground image~\cite{stoffregen2020reducing}. (d) Our synthesized dataset by graphics rendering, which not only reflects the real motions under correct scene geometries, but also produces accurate dense flow labels and events.}
    \label{fig:teaser}
\end{figure}
On the other hand, optical flow estimation predicts motions between two frames~\cite{sun2010secrets}, which is fundamental and important for many applications~\cite{vihlman2020optical,xu2019deep,capito2020optical}. In this work, we study the problem of estimating optical flow from event camera data,  instead of from RGB frames. Different from traditional images, events are sparse and are often integrated in short intervals as the input for the prediction. As such, early works can only estimate sparse flows at the location of events~\cite{benosman2012asynchronous}. Recent deep methods can estimate dense flows but with the help of images, either as the guidance~\cite{zhu2018ev} or as the additional inputs~\cite{lee2022Fusion-FlowNet,pan2020single}. Here, we tackle a hard version of the problem, where dense flows are predicted based purely on the event values $e$. One key issue is how to create high quality event-based optical flow dataset to train the network.      

Existing methods of event flow dataset creation can be classified into two types, 1) directly capturing from real event cameras~\cite{zhu2018multivehicle,zhu2018ev}; 2) moving foregrounds on top of a background image to create synthesized flow motions~\cite{wan2022DCEI,stoffregen2020reducing} and apply frame interpolation~\cite{gehrig2020video} to create events. For the first type, the  ground-truth (GT) flow labels need to be calculated based on gray images acquired along with the event data. However, the optical flow estimations cannot be perfectly accurate~\cite{teed2020raft,sui2022craft,luo2022kpa,jiang2021gma,xu2022gmflow}, leading to the inaccuracy of GT labels. To alleviate the problem, additional depth sensors, such as LIDAR, have been introduced~\cite{zhu2018multivehicle}. The flow labels can be calculated accurately when the depth values of LIDAR scans are available. However, LIDAR scans are sparse, and so do the flow labels, which are unfriendly for dense optical flow learning. Fig.~\ref{fig:teaser} (a) shows an example, LIDAR points on the ground are sparse. Moreover, some thin objects are often missing, as indicated by the red box in Fig.~\ref{fig:teaser} (a).

For the second category, the flow labels are created by moving foreground objects on top of a background image, similar to flying chairs~\cite{dosovitskiy2015flownet} or flying things~\cite{mayer2016large}. In this way, the flow labels are dense and accurate. To create events, intermediate frames are interpolated~\cite{jiang2018super}. However, the frame interpolation is inaccurate due to scene depth disparities, where the occluded pixels cannot be interpolated correctly, leading to erroneous event values in these regions. To match high framerate of events, the large number of interpolated frames makes the problem even worse. Fig.~\ref{fig:teaser} (b) shows an example, where the events are incorrect at the occluded chairs. Moreover, the motions are artificial, further decreasing the realism of the dataset (Fig.~\ref{fig:teaser} (b) and (c)). 

In this work, we create an event-flow dataset from synthetic 3D scenes by graphics rendering (Fig.~\ref{fig:teaser} (d)). While there is a domain gap between rendered and real images, this gap is empirically found insignificant in
event camera based classification~\cite{sironi2018hats} and segmentation~\cite{gehrig2020video} tasks. 
As noted by these works, models trained on synthetic events work very well for real event data. Because events contain only positive and negative polarities, no image appearances are involved. To this end, we propose a \textbf{M}ulti-\textbf{D}ensity \textbf{R}endered (MDR) event optical flow dataset, created by Blender on indoor and outdoor scenes with accurate events and flow labels. In addition, we design an \textbf{A}daptive \textbf{D}ensity \textbf{M}odule (ADM) based on MDR, which can adjust the densities of events, one of the most important factors for event-based tasks but has been largely overlooked. 

\begin{figure*}[!t]
    \centering
    \includegraphics[width=0.95\linewidth]{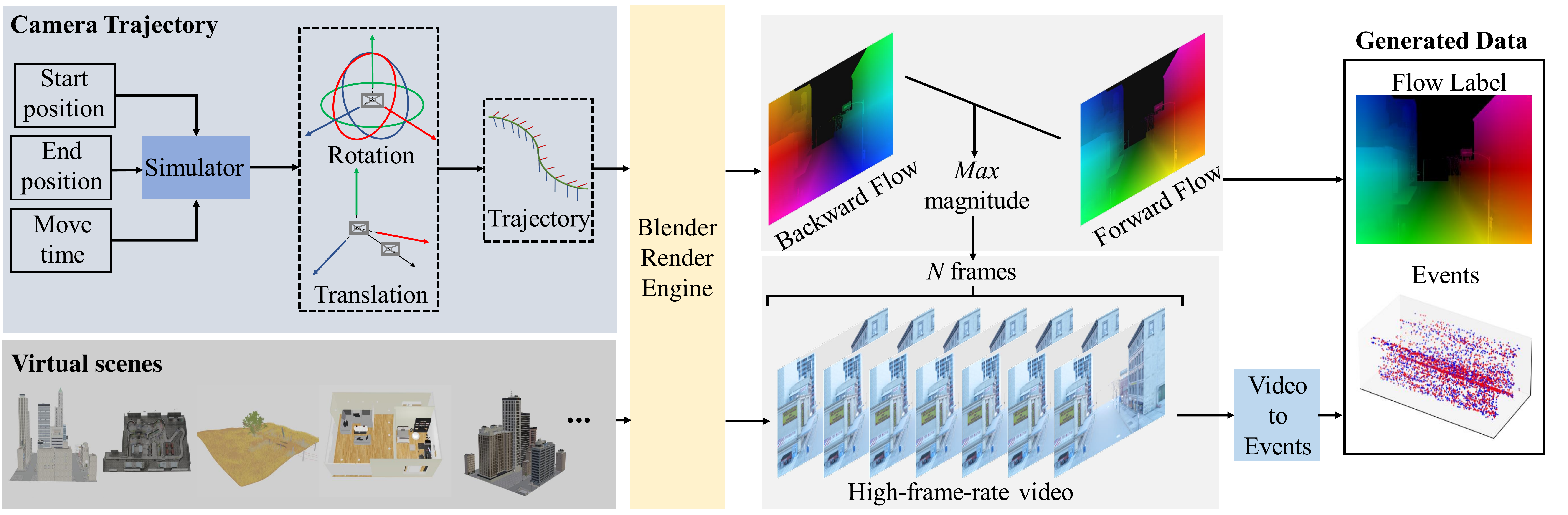}
    \caption{Our data generation pipeline. Given 3D scenes in graphics engine, we generate camera trajectories and render high-frame-rate videos with forward and backward optical flow labels. Then, we build the event optical flow dataset by generate events using the videos.}
    \label{fig:data_pipeline}
\end{figure*}

Specifically, our MDR dataset contains $80,000$ samples from $50$ virtual scenes. Each data sample is created by first rendering two frames and obtaining the GT flow  labels directly from the engine. Then, we render $15\sim60$ frames in-between based on the flow magnitude. The events are created by thresholding log intensities and recording the timestamp for each spatial location. The density of events can be controlled by the threshold values. The ADM is designed as a plugin module, which further consists of two sub-modules, multi-density changer (MDC) and multi-density selector (MDS), where the MDC adjusts the density globally while the MDS picks the best one for every spatial location. Experiments show that previous event-flow methods, when trained on our MDR dataset, can improve their performances. Moreover, we train several recent representative flow pipelines, such as FlowFormer~\cite{huang2022flowformer}, KPA-Flow~\cite{luo2022kpa}, GMA~\cite{jiang2021gma} and SKFlow~\cite{sun2022skflow}, on our MDR dataset. When equipped with our ADM module, the performances can increase consistently. 

Our contributions are summarized as:
\begin{itemize}
    \item 
    A rendered event-flow dataset MDR, with 80,000 samples created on 53 virtual scenes, which possess physically correct accurate events and flow label pairs, covering a wide range of densities. 
    
    \item
    An adaptive density module (ADM), which is a plug-and-play module for handling varying event densities.
    
    \item
    We achieve state-of-the-art performances. Our MDR can improve the quality of previous event-flow methods. Various optical flow pipelines when adapted to the event-flow task, can benefit from our ADM module.
\end{itemize}
\section{Related Work}
\subsection{Image-based Optical Flow}
Optical flow estimates per-pixel motion between two frames according to photo consistency. Traditional approaches minimize energies, leveraging both feature similarities and motion smoothness~\cite{fortun2015optical}. Deep methods train the networks that take two frames as input and directly output dense flow motions. Recent deep methods design different pipelines~\cite{dosovitskiy2015flownet,sun2018pwc,teed2020raft,jiang2021gma} as well as learning modules~\cite{luo2021upflow,liu2021oiflow,luo2022kpa,luo2022learning} for performance improvements. The training often requires large labeled datasets, which can be synthesized by moving a foreground on top of a background image, such as FlyingChairs~\cite{dosovitskiy2015flownet}, and autoflow~\cite{sun2021autoflow}, or rendered from graphics such as SinTel~\cite{butler2012naturalistic} and FlyingThings~\cite{mayer2016large}, or created directly from real videos~\cite{han2022realflow}. In this work, we use computer graphics techniques to render accurate and physically correct event and flow values.

\subsection{Event-based Optical Flow}
Benosman~\emph{et al.}~\cite{benosman2012asynchronous} first proposed to estimate optical flow from events, which can only estimate sparse flows at the location of event values. Recent deep methods can estimate dense optical flows. EV-FlowNet~\cite{zhu2018ev} learns the event and flow labels in a self-supervised manner, which minimizes the photometric distances of grey images acquired by DAVIS~\cite{brandli2014240}. Different event representations were explored, e.g., EST~\cite{gehrig2019EST} and Matrix-LSTM~\cite{cannici2020Matrix-LSTM}, with various network structures, such as SpikeFlowNet~\cite{lee2020spikeflownet}, LIF-EV-FlowNet~\cite{hagenaars2021LIF-EV-FlowNet}, STE-FlowNet~\cite{ding2022STE-FLOWNET}, Li~\emph{et al.}~\cite{li2021lightweight}, and E-RAFT~\cite{gehrig2021raft}. Some works take both events and images as input for the flow estimation~\cite{lee2022Fusion-FlowNet,pan2020single}. In general, dense flows are more desirable than sparse ones but are more difficult to train. Normally, regions with events can produce more accurate flows than empty regions where no events are triggered. Moreover, supervised training can produce better results than unsupervised ones, as long as the training dataset can provide sufficient event-flow guidance.    

\subsection{Event Dataset}
The applications to the event camera dataset were first explored in the context of classification~\cite{neil2016phased,bi2019graph}. Early works generate events simply by applying a threshold on the image difference~\cite{kaiser2016towards}. Frame interpolation is often adopted for high framerate~\cite{gehrig2020video}. The synthesized events often contain inaccurate timestamps. The DAVIS event camera can directly capture both images and events~\cite{brandli2014240}, based on which two driving datasets are captured, DDD17~\cite{binas2017ddd17} and MVSEC~\cite{zhu2018multivehicle}. With respect to the event flow dataset, EV-Flownet~\cite{zhu2018ev} calculated sparse flow labels from LIDAR depth based on MVSEC~\cite{zhu2018multivehicle}. Wan~\emph{et al.}~\cite{wan2022DCEI} and Stoffregen~\emph{et al.}~\cite{stoffregen2020reducing} created the foreground motions and interpolated the intermediate frames for events. The real captured dataset can only provide sparse labels while the synthesized ones contain inaccurate events. In this work, we propose to render a physically correct event dense flow dataset.

\section{Method}

\subsection{The event-based optical flow dataset}\label{sec: method: dataset}
In order to create a realistic event dataset for optical flow learning, we propose to employ a graphics engine with 3D scenes for data generation. Given 3D scenes, we first define camera trajectories, according to which we generate optical flow labels for timestamps at $60$ FPS and $15$ FPS. Then we render high-frame-rate videos based on the motion magnitude of the optical flow label between two timestamps. Finally, we generate event streams by rendering high-frame-rate videos and simulating the event trigger mechanism in the event camera using the v2e toolbox~\cite{hu2021v2e}. The overview of our data generation pipeline is shown in Fig.~\ref{fig:data_pipeline}.

\begin{figure}[!t]
    \centering
    \includegraphics[width=1.0\linewidth]{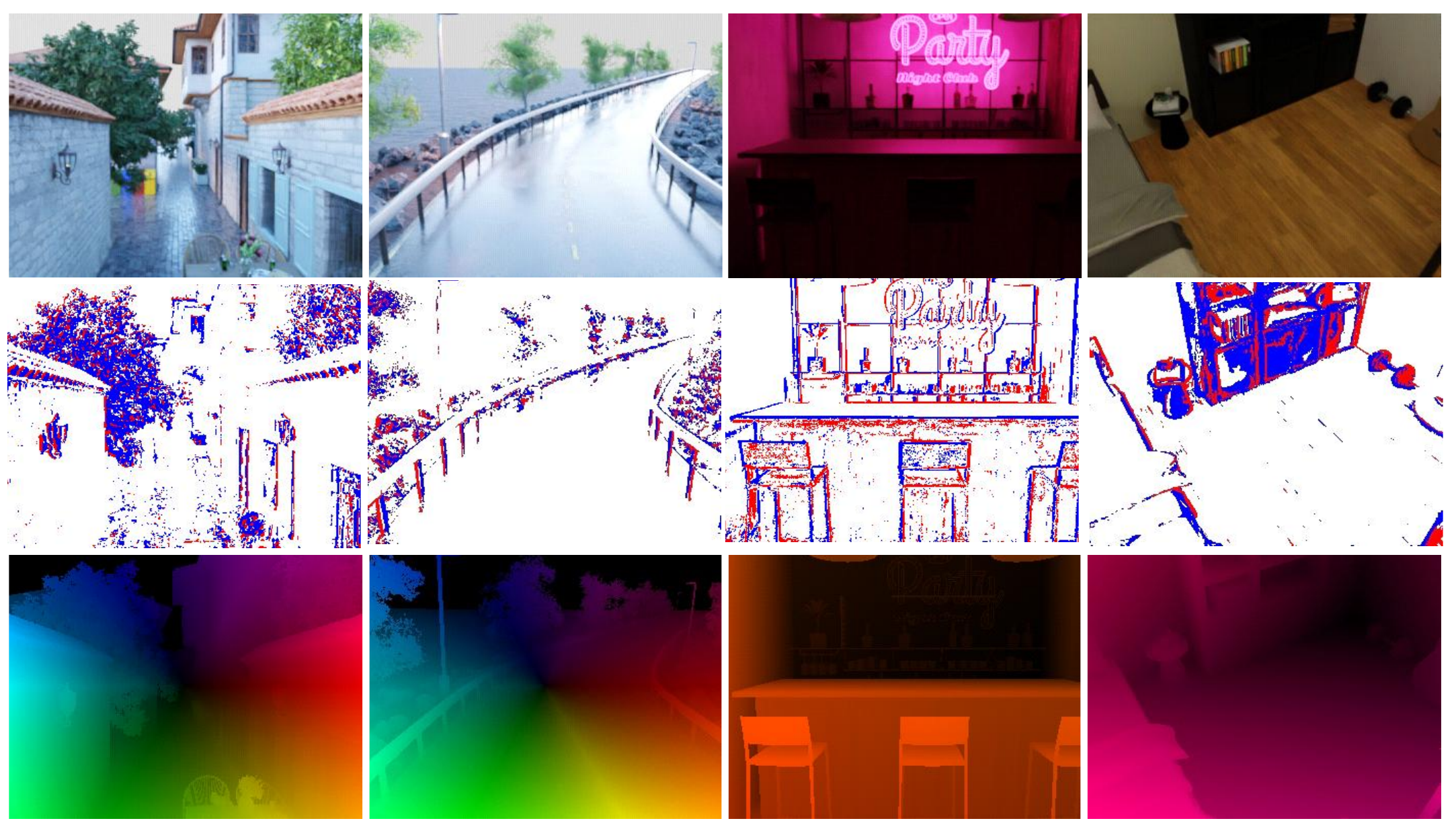}
    \caption{Examples of our MDR training set. Each row shows images, events and flow labels from top to bottom.}
    \label{fig:ourdatasets}
\end{figure}

 \paragraph{Virtual Scenes.}
 To ensure that the generated event dataset has the correct scene structure, we utilize a variety of indoor and outdoor 3D scenes, including cities, streets, forests, ports, beaches, living rooms, bedrooms, bathrooms, kitchens, and parking lots. Totally, we obtain $53$ virtual 3D scenes ($31$ indoor and $22$ outdoor) that simulate real-world environments. Some examples are shown in Fig.~\ref{fig:ourdatasets}.
 
 
 
 \paragraph{Camera Trajectory.}\label{sec: method: dataset: camera trajectory}
 Given a 3D scene model, we first generate the 3D camera trajectory using PyBullet~\cite{coumans2016pybullet}, an open-source physics engine, to ensure that the camera does not pass through the inside of the objects and out of the effective visible region of the scene during the motion. 
 After setting the start position, end position, and moving speed of the camera trajectory, we randomly add translation and rotation motions to create a smooth curve function $\Gamma(t)$ that outputs the location and pose $P(t)=[x(t),y(t),z(t), r(t)]^T$. 
 
 \paragraph{High-frame-rate Video and Optical Flow.}\label{sec: method: dataset: optical flow and video}
 After camera trajectory generation, we use the graphics engine to render a sequence of images $I(\bm{u},t_i)$, where $\bm{u}=(x,y)$ is the pixel coordinates and $t_i$ is the timestamp. We extract the forward and backward optical flow labels between every two timestamps ($\bm{F}_{t_i \to t_j}$, $\bm{F}_{t_j \to t_i}$). Then we need to generate the event data to construct the event optical flow dataset. Here, we render high-frame-rate videos $ \{ I(\bm{u},\tau) \}, \tau \in [t_i,t_j] $ between timestamps $t_i$ and $t_j$ for events generation according to the camera trajectory $\Gamma(t)$. 

 Inspired by ESIM~\cite{rebecq2018esim}, we adopt an adaptive sampling strategy to sample camera locations from the camera trajectory for interval $t_i$ to $t_j$, so that the largest displacement of all pixels between two successive rendered frames ($I(\bm{u},\tau_k)$, $I(\bm{u},\tau_{k+1})$) is under 1 pixel, we define the sampling time interval $\Delta \tau_{k}$ as follows:
 \begin{equation}
\begin{aligned}
\Delta \tau_{k} &= \tau_{k+1} - \tau_{k} \\
&= (\max_{\bm{u}} \max \{ \left \| \bm{F}_{{\tau_{k-1}} \to {\tau_{k}}} \right \| , \left \| \bm{F}_{\tau_{k} \to \tau_{k-1}} \right \|\})^{-1},
\end{aligned}
\label{sampling}
\end{equation}
where $\left | F \right| = \max_{\bm{u}} \max \{ \left \| \bm{F}_{{\tau_{k-1}} \to {\tau_{k}}} \right \| , \left \| \bm{F}_{{\tau_{k}} \to {\tau_{k-1}}} \right \|\}$ is the maximum magnitude of the motion field between images $I(\bm{u},\tau_{k-1})$ and $I(\bm{u},\tau_{k})$. 

 \paragraph{Event Generation from High-frame-rate Video.}
 Given a high-frame-rate video $ \{ I(\bm{u},\tau) \}, \tau \in [t_i,t_j] $ between timestamps $t_i$ and $t_j$, we next generate event stream by simulating the event trigger mechanism. Similar to \cite{rebecq2018esim} and \cite{gehrig2020video}, we use linear interpolation to approximate the continuous intensity signal in time for each pixel between video frames. 
Events $\left \{ (\bm{u_e},t_e,p_e) \right \}$ are generated at each pixel $\bm{u_e}=(x_e,y_e)$ whenever the magnitude of the change in the log intensity values ($L(\bm{u_e},t_e) = \ln(I(\bm{u_e},t_e)$) exceeds the threshold $C$. This can be expressed as Eq.(\ref{interpolation1}) and  Eq.(\ref{interpolation2}):


 \begin{equation}
L(\bm{u_e},t_e+\Delta t_e) - L(\bm{u_e},t_e) \ge  p_e C,
\label{interpolation1}
\end{equation}
 \begin{equation}
t_{e} = t_{e-1} + \Delta \tau_{k} \frac{C}{\left | L(\bm{u_e},t_e+\Delta t_e) - L(\bm{u_e},t_e) \right | },
\label{interpolation2}
\end{equation}
where $t_{e-1}$ and $t_{e}$ are the timestamps of the last triggered event and the next triggered event respectively, $p_e \in \left \{ -1,+1  \right \} $ is the polarity of the triggered event. We define it as $E(t_{k},t_{k+1})$, witch is the sequence $ \{(\bm{u_e},t_e,p_e)^N, e \in [ 0,N ]\}$ with $N$ events between time $t_{k}$ and $t_{k+1}$.


\paragraph{Multi-Density Rendered Events Dataset.}
Using the above data generation method, we can generate data with different event densities by using different threshold values $C$. 
Since event stream is commonly first transformed into event representation~\cite{zhu2019unsupervised,stoffregen2020reducing,gehrig2021raft} and then fed into deep networks. In order to measure the amount of useful information carried by the event stream, we propose to calculate the density of the event stream using the percentage of valid pixels (pixels where at least one event is triggered) in the voxel representation:
 \begin{equation}
V(\bm{u_e}, b) = \sum_{e=0}^{N}p_{e} \max(0, 1-|b-\frac{t_e-t_0}{t_N-t_0}(B-1)|),
\label{reprentation}
\end{equation}
 \begin{equation}
D = \frac{1}{HW}\sum_{i=0}^{N} \varepsilon (\sum_{b=0}^{B}|V(\bm{u_e}, b)|),
\varepsilon(x)=\left\{\begin{matrix}1 , x>0\\0,x \le 0\end{matrix}\right.,
\label{density}
\end{equation} 
where $V(\bm{u_e}) \in \mathbb{R}^{B\times H \times W}$ is the voxel representation~\cite{zhu2019unsupervised} of the event stream $\left \{ (\bm{u_e},t_e,p_e)^N \right \}$ between $t_0$ and $t_N$, $b \in \left [ 0, B-1 \right ] $ indicates the temporal index , $B$ (typically set to 5) donates temporal bins and $D$ is the density of the input event representation $V$.
In practical applications, different event cameras may use different threshold values in different scenes, resulting in data with different event densities. Intuitively, event data with lower density is more difficult for optical flow estimation. In order to train models that can cover event data with various density, in this paper, we propose to adaptively normalize the density of the input events to a certain density representation for optical flow estimation, so as to increase the generalization ability of the network.


\subsection{Event-based Optical Flow Estimation}
\begin{figure}[!t]
    \centering
    \includegraphics[width=\linewidth]{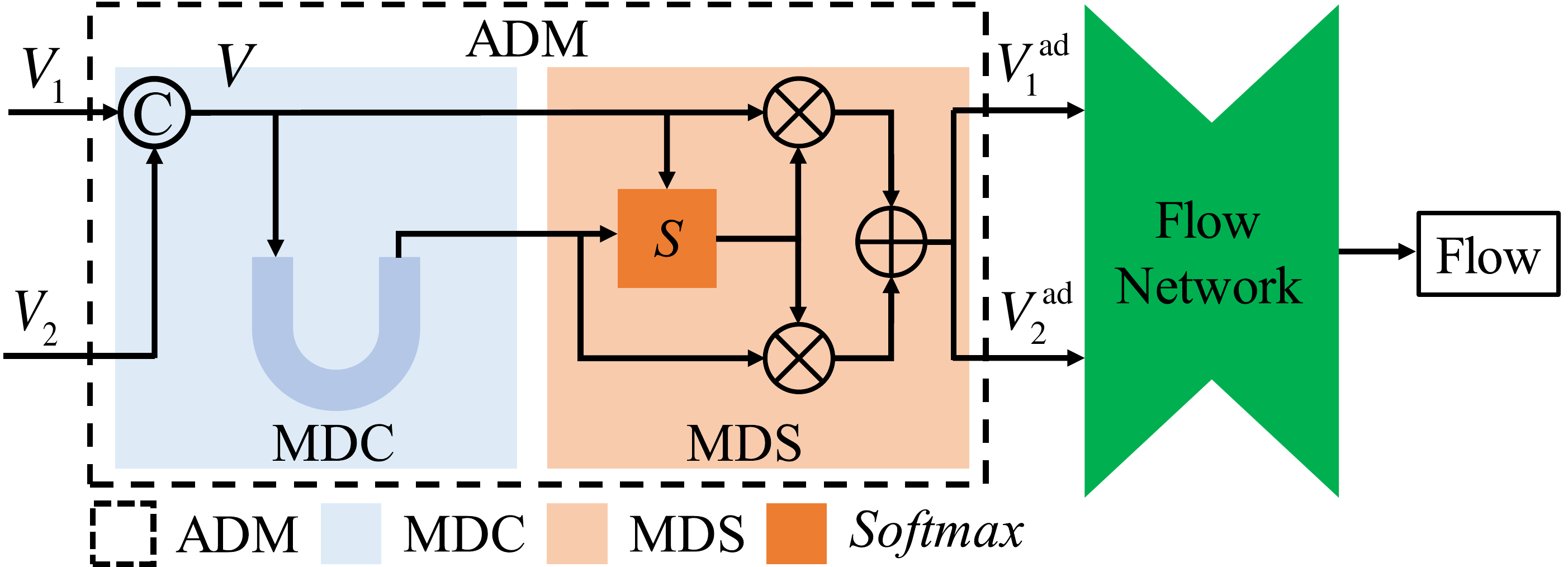}
    \caption{The structure of the proposed network. We design a plug-and-play Adaptive Density Module (ADM) to transform input event representations $V_1$ and $V_2$ into 
    $V_1^{\mathrm{ad}}$ and $V_2^{\mathrm{ad}}$ with suitable density for optical flow estimation. }
    \label{fig:networks}
\end{figure}

Event-based optical flow estimation involves predicting dense optical flow $\bm{F}_{k-1 \to k}$ from consecutive event sequences $E(t_{k-1},t_{k})$ and $E(t_{k},t_{k+1})$. In this paper, we find that networks perform better on event sequences with appropriate density than on those with excessively sparse or dense events, when given events from the same scene. Motivated by this, we propose a plug-and-play Adaptive Density Module (ADM) that normalizes the input event stream to a density suited for estimating optical flow. Our network architecture is shown in Fig.~\ref{fig:networks}, where the ADM transforms input event representations $V_1$ and $V_2$ to justified event representations $V_1^{\mathrm{ad}}$ and $V_2^{\mathrm{ad}}$, which are then used by an existing network structure to estimate optical flow. 



\paragraph{Adaptive Density Module.}
As shown in Fig.~\ref{fig:networks}, our ADM module consists of two sub-modules: the multi-density changer (MDC) module and the multi-density selector (MDS) module. The MDC module globally adjusts the density of the input event representations from multi-scale features, then the MDS module picks the best pixel-wise density for optical flow estimation.

\begin{figure}[!t]
    \centering
    \includegraphics[width=0.8\linewidth]{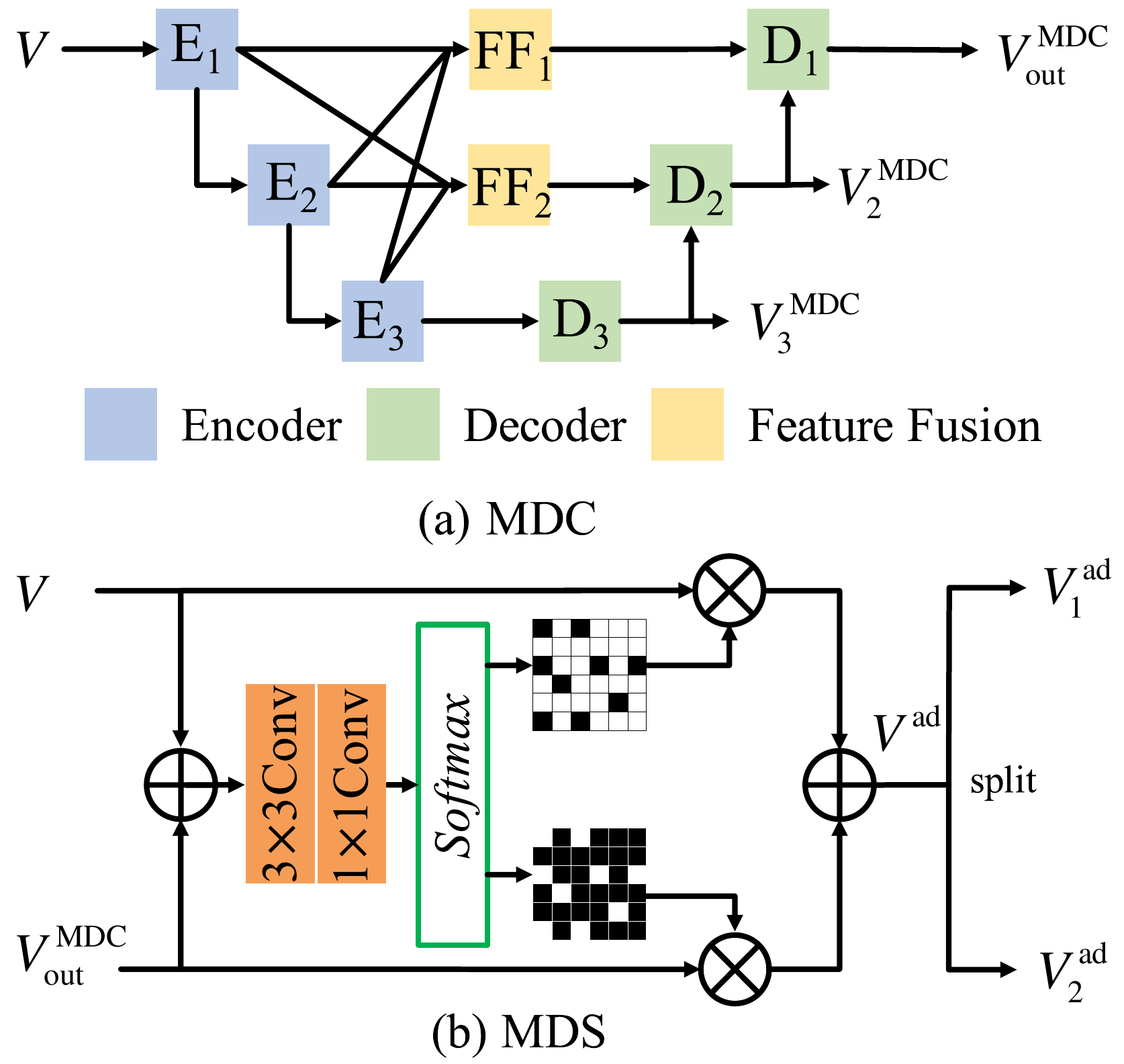}
    \caption{The detailed structure of sub-modules used in our proposed ADM model: (a) MDC, (b) MDS.}
    \label{fig:submodule}
\end{figure}

The MDC module adopts an encoder-decoder architecture with three levels, as illustrated in Fig.~\ref{fig:submodule}(a). To generate multiscale transformed representations $V_3^{\mathrm{MDC}}$, $V_2^{\mathrm{MDC}}$ and $V_1^{\mathrm{MDC}}$ (also noted as $V_{\mathrm{out}}^{\mathrm{MDC}}$) from the concatenated input event representations $V$, three encoding blocks are employed to extract multiple scale features, followed by three decoding blocks and two feature fusion blocks. It is worth noting that, to ensure the lightweightness of the entire module, we utilize only two $3\times 3$ and one $1\times 1$ convolutional layers in each encoding and decoding block.


To maintain the information in the input event representation and achieve density transformation, we adopt the MDS module for adaptive selection and fusion of $V_{\mathrm{out}}^{\mathrm{MDC}}$ and $V$, as depicted in Fig.~\ref{fig:submodule}(b). We first concatenate $V_{\mathrm{out}}^{\mathrm{MDC}}$ and $V$, and then use two convolutional layers to compare them and generate selection weights via softmax. Finally, we employ the selection weights to identify and fuse $V_{\mathrm{out}}^{\mathrm{MDC}}$ and $V$, producing the transformed event representation $V_1^{\mathrm{ad}}$ and $V_2^{\mathrm{ad}}$, which are fed into an existing flow network for optical flow estimation. In this paper, we use KPA-Flow~\cite{luo2022kpa} for optical flow estimation by default.

\paragraph{Loss Function.}
Based on our MDR dataset, we use the event representation with moderate density as the ground truth (noted as $V^\mathrm{GT}$) to train our ADM module. 
For the MDC module, we use a multi-scale loss as follows:
 \begin{equation}
L_{\mathrm{MDC}}=\sum_{k=1}^{3}\sqrt{(V_k^\mathrm{MDC}-V_k^\mathrm{GT})^2+\xi^2},
\label{MDCloss}
\end{equation} 
where $\xi=10^{-3}$ is a constant value, $V_k^\mathrm{MDC}$ is the output of the $k$-th level in MDC, and $V_k^\mathrm{GT}$ is downsampled from $V^\mathrm{GT}$ to match the spatial size.

For the MDS module, we use the distance between the density of $V^\mathrm{ad}$ and $V^\mathrm{GT}$ as the guidance:
 \begin{equation}
L_{\mathrm{MDS}}=\left \| D(V^\mathrm{ad}) - D(V^\mathrm{GT}) \right \|_1,
\label{MDSloss}
\end{equation} 
where $D$ means to calculate the density as in Eq.(~\ref{density}).

For the flow network, we use L1 loss (denoted as $L_{\mathrm{Flow}}$ ) between flow prediction and ground truth as the guidance. 

The final loss function for training the whole pipeline in Fig.~\ref{fig:networks} is determined as follows:
 \begin{equation}
L_{total} = \lambda_1 L_{\mathrm{MDC}} + \lambda_2 L_{\mathrm{MDS}} + L_{\mathrm{Flow}},
\label{totalloss}
\end{equation} 
where we empirically set $\lambda_1 = 0.1$ and $\lambda_2 = 10$.


\begin{table*}[t]
	\centering
	\resizebox*{0.9 \linewidth}{!}{
		\begin{tabular}
			{   >{\arraybackslash}p{3.4cm}| 
                >{\centering\arraybackslash}p{1.0cm}| 
				>{\centering\arraybackslash}p{0.8cm}| 
				>{\centering\arraybackslash}p{0.7cm} 
				>{\centering\arraybackslash}p{0.7cm}| 
				>{\centering\arraybackslash}p{0.7cm} 
				>{\centering\arraybackslash}p{0.7cm}| 
    			>{\centering\arraybackslash}p{0.7cm} 
				>{\centering\arraybackslash}p{0.7cm}| 
    			>{\centering\arraybackslash}p{0.7cm} 
				>{\centering\arraybackslash}p{0.7cm}| 
        		>{\centering\arraybackslash}p{0.7cm} 
				>{\centering\arraybackslash}p{0.7cm} 
			}
			\hline
             \multirow{2}{*}{Method ($dt=1$)} & Train&Train& \multicolumn{2}{c|}{indoor flying1} &  \multicolumn{2}{c|}{indoor flying2} &  \multicolumn{2}{c|}{indoor flying3} &  \multicolumn{2}{c|}{outdoor day1} &  \multicolumn{2}{c}{Avg} \\
             & D.Type&D.Set& EPE   & \%Out & EPE   & \%Out & EPE   & \%Out & EPE   & \%Out & EPE   & \%Out \\
             \hline
            EST$_S$~\cite{gehrig2019EST} & $\rm{E}$      & M & 0.97  & 0.91  & 1.38  & 8.20  & 1.43  & 6.47  & -     & -     & 1.26  & 5.19  \\
    EV-Flownet$_S$~\cite{zhu2018ev} & $\rm{I_{1},I_{2},E}$ & M & 1.03  & 2.20  & 1.72  & 15.1  & 1.53  & 11.9  & 0.49  & 0.20  & 1.19  & 7.35  \\
    Deng et al.$_S$~\cite{deng2021learning} & $\rm{E}$      & M & 0.89  & 0.66  & 1.31  & 6.44  & 1.13  & 3.53  & -     & -     & 1.11  & 3.54  \\
    Paredes et al.$_S$~\cite{paredes2021back} & $\rm{E}$      & M & 0.79  & 1.20  & 1.40  & 10.9  & 1.18  & 7.40  & 0.92  & 5.40  & 1.07  & 6.22  \\
    Matrix-LSTM$_S$~\cite{cannici2020Matrix-LSTM} & $\rm{I_{1},I_{2},E}$ & M & 0.82  & 0.53  & 1.19  & 5.59  & 1.08  & 4.81  & -     & -   & 1.03  & 3.64  \\
    LIF-EV-FlowNet$_S$~\cite{hagenaars2021LIF-EV-FlowNet} & $\rm{E}$      & FPV   & 0.71  & 1.41  & 1.44  & 12.8  & 1.16  & 9.11  & 0.53  & 0.33  & 0.96  & 5.90  \\
    Spike-FlowNet$_S$~\cite{lee2020spikeflownet} & $\rm{I_{1},I_{2},E}$ & M & 0.84  & -     & 1.28  & -     & 1.11  & -     & 0.49  & -     & 0.93  & - \\
    Fusion-FlowNet$_D$~\cite{lee2022Fusion-FlowNet} & $\rm{I_{1},I_{2},E}$ & M & 0.62  & -     & 0.89  & -     & 0.85  & -     & 1.02  & -     & 0.84  & - \\
    Fusion-FlowNet$_S$~\cite{lee2022Fusion-FlowNet} & $\rm{I_{1},I_{2},E}$ & M & 0.56  & -     & 0.95  & -     & 0.76  & -     & 0.59  & -     & 0.71  & - \\
    Zhu et al.$_S$~\cite{zhu2019unsupervised} & $\rm{E}$      & M & 0.58  & \textcolor{red}{0.00}  & 1.02  & 4.00  & 0.87  & 3.00  & \textcolor{red}{ 0.32}  & \textcolor{blue}{ 0.00 }  & 0.69  & 1.75  \\
    DCEIFlow$_D$~\cite{wan2022DCEI} & $\rm{I_{1},I_{2},E}$ & C2    & 0.56  & 0.28  & \textcolor{blue}{ 0.64 }  & \textcolor{red}{ 0.16 } & 0.57  & 0.12  & 0.91  & 0.71  & 0.67  & 0.31  \\
    DCEIFlow$_S$~\cite{wan2022DCEI}& $\rm{I_{1},I_{2},E}$ & C2    & 0.57  & 0.30 & 0.70  & \textcolor{blue}{ 0.30 }  & 0.58  & 0.15  & 0.74  & 0.29  & 0.64  & \textcolor{blue}{ 0.26 } \\
    Stoffregen et al.$_S$~\cite{stoffregen2020reducing} & $\rm{E}$      & ESIM  & 0.56  & 1.00  & 0.66  & 1.00  & 0.59  & 1.00  & 0.68  & 0.99  & 0.62  & 0.99  \\
    STE-FlowNet$_S$~\cite{ding2022STE-FLOWNET}& $\rm{I_{1},I_{2},E}$ & M & 0.57  & \textcolor{blue}{ 0.10 }  & 0.79  & 1.60  & 0.72  & 1.30  & 0.42  & 0.00  & 0.62  & 0.75  \\
    \bf ADM-Flow$_D$(ours) & $\rm{E}$      & MDR   & \textcolor{red}{ 0.48 } & 0.11  & \textcolor{red}{ 0.56 } & 0.40  & \textcolor{red}{ 0.47 } & \textcolor{red}{ 0.02 } & 0.52  & 0.00 & \textcolor{red}{ 0.51 } & \textcolor{red}{ 0.14 }   \\
    \bf ADM-Flow$_S$(ours) & $\rm{E}$      & MDR   & \textcolor{blue}{ 0.52 } & 0.14  & 0.68  & 1.18  & \textcolor{blue}{ 0.52 }  & \textcolor{blue}{0.04}  & \textcolor{blue}{ 0.41 }  & \textcolor{red}{ 0.00 } & \textcolor{blue}{ 0.53 }  & 0.34  \\
	\hline
    \multirow{2}{*}{Method ($dt=4$)} & Train&Train& \multicolumn{2}{c|}{indoor flying1} &  \multicolumn{2}{c|}{indoor flying2} &  \multicolumn{2}{c|}{indoor flying3} &  \multicolumn{2}{c|}{outdoor day1} &  \multicolumn{2}{c}{Avg} \\
 & D.Type&D.Set& EPE   & \%Out & EPE   & \%Out & EPE   & \%Out & EPE   & \%Out & EPE   & \%Out \\
 \hline
 LIF-EV-FlowNet$_S$~\cite{hagenaars2021LIF-EV-FlowNet} & $\rm{E}$      & FPV   & 2.63  & 29.6  & 4.93  & 51.1  & 3.88  & 41.5  & 2.02  & 18.9  & 3.36  & 35.3  \\
    EV-Flownet$_S$~\cite{zhu2018ev} & $\rm{I_{1},I_{2},E}$ & M & 2.25  & 24.7  & 4.05  & 45.3  & 3.45  & 39.7  & 1.23  & \textcolor{blue}{ 7.30 } & 2.74  & 29.3  \\
    Zhu et al.$_S$~\cite{zhu2019unsupervised} & $\rm{E}$      & M & 2.18  & 24.2  & 3.85  & 46.8  & 3.18  & 47.8  & 1.30  & 9.70  & 2.62  & 32.1  \\
    Spike-FlowNet$_S$~\cite{lee2020spikeflownet} & $\rm{I_{1},I_{2},E}$ & M & 2.24  & -     & 3.83  & -     & 3.18  & -     & \textcolor{blue}{ 1.09 }  & -     & 2.58  & - \\
    Fusion-FlowNet$_D$~\cite{lee2022Fusion-FlowNet} & $\rm{I_{1},I_{2},E}$ & M & 1.81  & -     & 2.90  & -     & 2.46  & -     & 3.06  & -     & 2.55  & - \\
    Fusion-FlowNet$_S$~\cite{lee2022Fusion-FlowNet}& $\rm{I_{1},I_{2},E}$ & M & 1.68  & -     & 3.24  & -     & 2.43  & -     & 1.17  & -     & 2.13  & - \\
    STE-FlowNet$_S$~\cite{ding2022STE-FLOWNET} & $\rm{I_{1},I_{2},E}$ & M & 1.77  & 14.7  & 2.52  & 26.1  & 2.23  & 22.1  & \textcolor{red}{ 0.99 } & \textcolor{red}{ 3.90 } & 1.87  & 16.7  \\
    DCEIFlow$_D$~\cite{wan2022DCEI} & $\rm{I_{1},I_{2},E}$ & C2    & 1.49  & 8.14  & 1.97  & 17.4  & 1.69  & 12.3  & 1.87  & 19.1  & 1.75  & 14.24  \\
    DCEIFlow$_S$~\cite{wan2022DCEI} & $\rm{I_{1},I_{2},E}$ & C2    & 1.52  & 8.79  & 2.21  & 22.1  & 1.74  & 13.3  & 1.37  & 8.54  & 1.71  & 13.2  \\
    \bf ADM-Flow$_D$(ours) & $\rm{E}$      & MDR    & \textcolor{red}{ 1.39 } & \textcolor{red}{ 7.33 } & \textcolor{red}{ 1.63 } & \textcolor{red}{ 11.5 } & \textcolor{red}{ 1.51 } & \textcolor{red}{ 9.34 } & 1.91  & 19.2  & \textcolor{blue}{ 1.61 }  & \textcolor{blue}{ 11.8 } \\
    \bf ADM-Flow$_S$(ours) & $\rm{E}$      & MDR   & \textcolor{blue}{ 1.42 }  & \textcolor{blue}{ 7.78 } & \textcolor{blue}{ 1.88 } & \textcolor{blue}{ 16.7 } & \textcolor{blue}{ 1.61 } & \textcolor{blue}{ 11.4 }  & 1.51  & 10.23  & \textcolor{red}{ 1.60 } & \textcolor{red}{ 11.5 } \\
 \hline
	\end{tabular}}
    \caption{Quantitative comparison of our method with previous methods on the MVSEC dataset \cite{zhu2018multivehicle}.
    Subscripts $S$ and $D$ donate the \emph{sparse} and \emph{dense} evaluation, respectively. We mark the best results in \textcolor{red}{red} and the second best results in \textcolor{blue}{blue}.}
	\label{tab:mvsec_result}
\end{table*}

\section{Experiments}
\subsection{Datasets}
\noindent{\bf MVSEC:} The MVSEC dataset \cite{zhu2018multivehicle} is a real-world dataset collected in indoor and outdoor scenarios with sparse optical flow labels. As a common setting, $28,542$ data pairs of the `outdoor day2' sequence are used as the train set, and $8,410$ data pairs of the other sequences are used as the validation set. The density ranges of MVSEC train set and validation set are $[0.0003, 0.47]$ and $[0.001, 0.31]$, respectively.

\noindent{\bf MDR:} We create the MDR dataset using the graphics engine blender. We use various 3D scenes for data generation. There are $80,000$ training samples and $6,000$ validation samples with accurate dense optical flow ground truth. 
Each sample has event sequences with different density produced by using different threshold $C$. By default, for training on MDR, we use the combination of all these samples with different densities to train flow networks. For the learning of our ADM module, we choose events with density between 0.45 and 0.55 as the lable for $L_\mathrm{MDC}$ and $L_\mathrm{MDS}$.

Previous methods also train their models on other datasets, including synthetic datasets C2~\cite{wan2022DCEI} and ESIM~\cite{stoffregen2020reducing}, as well as the real-world captured dataset FPV~\cite{Delmerico19icra}. We also compare with tham on the validation set of the MVSEC dataset. Following~\cite{ding2022STE-FLOWNET}, We train and evaluate all the methods using event sequences sampled at 60 Hz (denoted as $dt=1$) and 15 Hz (denoted as $dt=4$).

\subsection{Implementation details}
We use PyTorch to implement our method and train all networks using the same setting. The networks are trained with the AdamW optimizer, with a batch size of 6 and learning rate of $4 \times10^{-4}$ for 150k iterations. Since the MVSEC dataset lacks multiple density event streams required for the learning of our ADM module, we disable $L_{\mathrm{MDC}}$ and $L_{\mathrm{MDS}}$ when training on the MVSEC dataset. For evaluation, we use the average End-point Error (EPE) and the percentage of points with EPE greater than 3 pixels and 5\% of the ground truth flow magnitude, denoted as \%Out. We calculate errors at the pixel with valid flow annotation for \emph{dense} evaluation, and at the pixel with valid flow annotation and triggered at least one event for \emph{sparse} evaluation.



\subsection{Comparison with State-of-the-Arts} 
\begin{figure*}[!t]
    \centering
    \includegraphics[width=\linewidth]{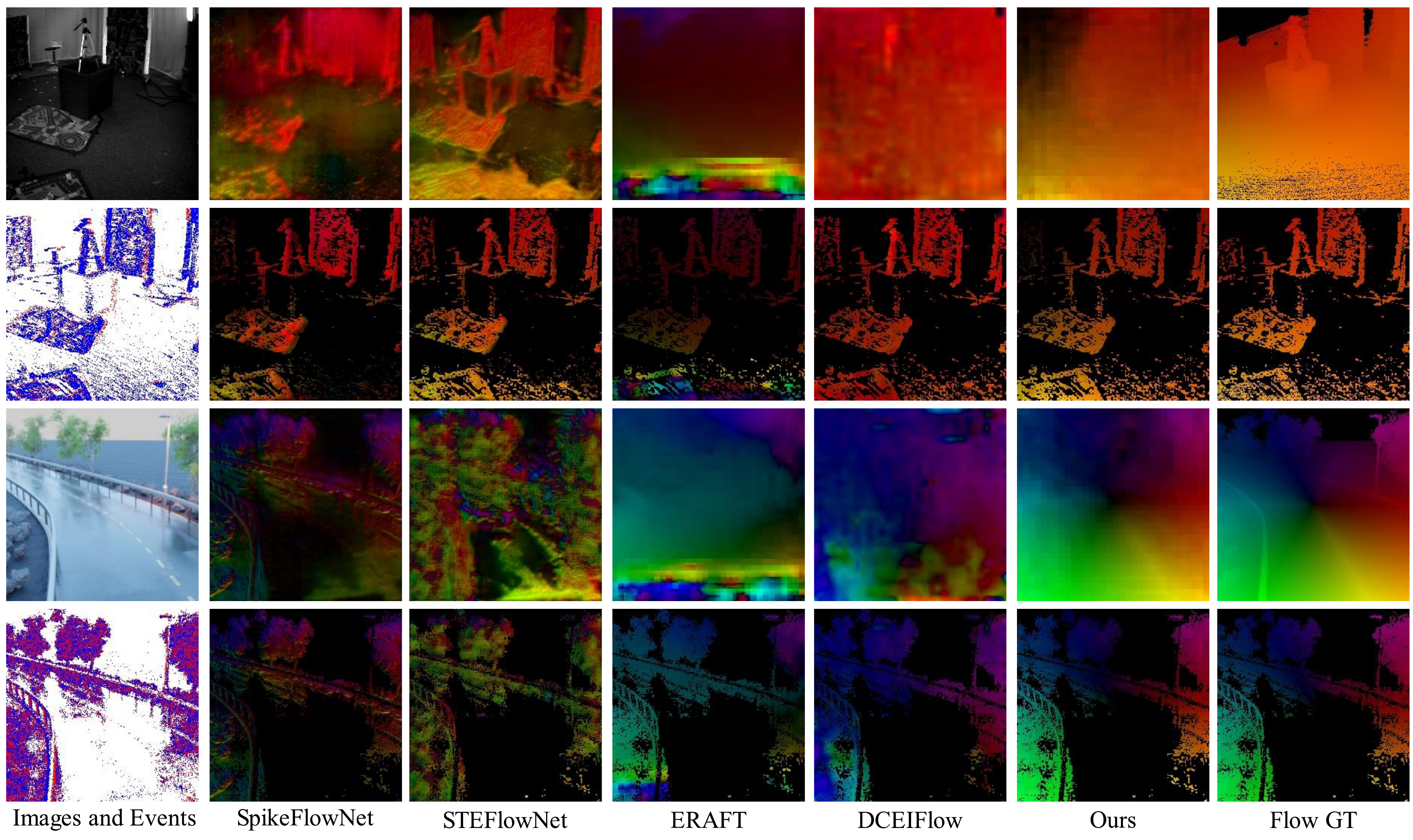}
    \caption{Qualitative comparisons compared with existing event-based methods. Row 1 and 2 are from MVSEC, whereas row 3 and 4 are from MDR. Row 1 and 3 visualize the dense predictions, whereas row 2 and 4 show the sparse.}
    \label{fig:comparison}
\end{figure*}

\paragraph{Results on MVSEC.} 
In Table~\ref{tab:mvsec_result}, we compare our model trained on the MDR train set with previous methods on the MVSEC evaluation set, and report detailed results for each sequence. We provide information on the data types (Train D.Type) and the training sets (Train D.Set) used in the training process for each method. Specifically, `$\rm{I_{1},I_{2},E}$' indicates that both image data and event data are used in the training and inference processes of the model, while `$\rm{E}$' indicates that only event data is used. As shown in Table~\ref{tab:mvsec_result}, our model trained on the MDR dataset achieves state-of-the-art performance for both EPE and outlier metrics in settings of $dt=1$ and $dt=4$. Notably, our method demonstrates a 23.9\% improvement (reducing EPE from 0.67 to 0.51) for dense optical flow estimation in $dt=1$ settings, and an 8.0\% improvement (reducing EPE from 1.75 to 1.61) in $dt=4$ settings, surpassing previous methods. Qualitative comparison results are shown in row 1 and row 2 in Fig.~\ref{fig:comparison}.

\paragraph{Results on MDR.} 
In Table~\ref{tab:mdr_result}, we compare our method with previous methods for training on MVSEC dataset and testing on MDR dataset. We use different threshold $C$ to generate test data with different density ranges for evaluation. For the average EPE error of dense optical flow estimation, our model obtains the best result, which is a 17.1\% improvement (reducing EPE from 0.70 to 0.58) in $dt=1$ settings, and a 12.0\% improvement (reducing EPE from 1.67 to 1.47) in $dt=4$ settings. We also show some qualitative comparison results in row 3 and row 4 in Fig.~\ref{fig:comparison}.


\paragraph{Analysis of the MDR dataset.}  
To demonstrate the effectiveness of our proposed MDR dataset, we train several optical flow networks~\cite{sun2018pwc,gehrig2021raft,zhao2022GMFlowNet,huang2022flowformer,sun2022skflow,jiang2021gma,luo2022kpa} on both MDR and MVSEC train sets using identical training settings. For training, we use a combination of data samples from the MDR dataset with a density range of $[0.09, 0.69]$. We evaluate the trained networks on the MVSEC validation set, and the results, presented in Table~\ref{tab:dataset_comparison}, demonstrate that all networks trained on the MDR dataset outperform those trained on the MVSEC dataset.

\begin{table}[t]
	\centering
	\resizebox*{0.99 \linewidth}{!}{
		\begin{tabular}
			{   >{\arraybackslash}p{2.9cm}| 
				>{\centering\arraybackslash}p{1.4cm}| 
				>{\centering\arraybackslash}p{1.4cm}| 
				>{\centering\arraybackslash}p{1.4cm}| 
				>{\centering\arraybackslash}p{1.4cm}| 
				>{\centering\arraybackslash}p{0.65cm} 
			}
\hline
\multirow{2}{*}{Method ($dt=1$)} &  \multicolumn{4}{c|}{density range} & \multirow{2}{*}{Avg}  \\
\cline{2-5}
&0.09-0.21&0.21-0.36&0.36-0.57&0.57-0.69&\\
\hline
Spike-Flow$_S$~\cite{lee2020spikeflownet} & 1.13   & 1.20   & 1.33    & 2.01    & 1.42    \\
STE-Flow$_S$~\cite{ding2022STE-FLOWNET} & 0.82   & 1.00    & 0.87    & 0.82    & 0.88   \\
E-RAFT$_D$~\cite{gehrig2021raft} & 0.93   & 0.85    & 0.69   & 0.92    & 0.85    \\
E-RAFT$_S$~\cite{gehrig2021raft} & 1.02    & 1.13   & 0.89    & 1.14    & 1.05    \\
DCEIFlow$_D$~\cite{wan2022DCEI} & \textcolor{blue}{0.79}   & 0.72    & 0.67    & \textcolor{blue}{0.62}    & 0.70   \\
DCEIFlow$_S$~\cite{wan2022DCEI} & 0.90    & 0.66    & 0.60    & 0.98   & 0.79    \\
\bf ADM-Flow$_D$(ours) & \textcolor{red}{0.73}  & \textcolor{blue}{0.53}  & \textcolor{blue}{0.51}  &  \textcolor{red}{0.56}  &  \textcolor{red}{0.58}  \\
\bf ADM-Flow$_S$(ours) & 0.96   &  \textcolor{red}{0.47}   &  \textcolor{red}{0.47}   & 0.63    & \textcolor{blue}{0.63}  \\
\hline
\multirow{2}{*}{Method ($dt=4$)} &  \multicolumn{4}{c|}{density range} & \multirow{2}{*}{Avg}  \\
\cline{2-5}
&0.09-0.21&0.21-0.36&0.36-0.57&0.57-0.69&\\
\hline
Spike-Flow$_S$~\cite{lee2020spikeflownet} & 3.95    & 1.96    & 2.09  & 2.87    & 2.72   \\
STE-Flow$_S$~\cite{ding2022STE-FLOWNET} & 2.71   & 2.17    & 1.72   & \textcolor{red}{1.73}   & 2.08    \\
DCEIFlow$_D$~\cite{wan2022DCEI} &  \textcolor{red}{1.72}    & 1.66   & 1.16    & 2.14    & \textcolor{blue}{1.67}    \\
DCEIFlow$_S$~\cite{wan2022DCEI} & 2.87   & 1.35    & 1.16    & 2.24   & 1.90    \\
\bf ADM-Flow$_D$(ours) & \textcolor{blue}{1.89}  &  \textcolor{red}{1.25}   &  \textcolor{red}{0.98}  &  \textcolor{blue}{1.75}  &  \textcolor{red}{1.47} \\
\bf ADM-Flow$_S$(ours) & 2.54   & \textcolor{blue}{1.27}   & \textcolor{blue}{1.09}  & 2.17   & 1.77   \\
\hline
	\end{tabular}}
    \caption{Quantitative evaluation on our MDR dataset. The methods in table are all trained on MVSEC dataset. $S$ and $D$ donate the \emph{sparse} and \emph{dense} evaluation, respectively. We use EPE as the evaluation metric. }
	\label{tab:mdr_result}
\end{table}

\subsection{Ablation study.} 

\begin{figure}[t]
    \centering
    \includegraphics[width=\linewidth]{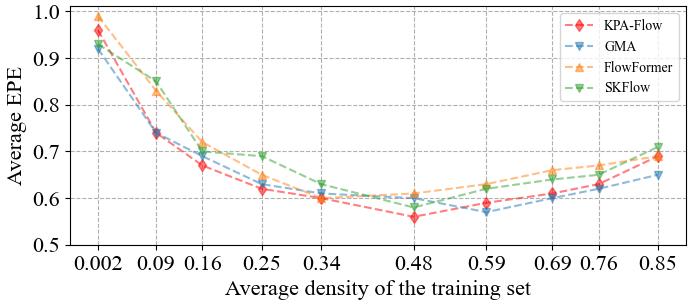}
    \caption{The performance of some supervised optical flow networks with different densities of the training set. X-axis is the average density of events in the training set, and y-axis is their average EPE in the test set of MVSEC.}
    \label{fig:density}
\end{figure}

\paragraph{Training with different densities.\label{densities_comparison}}
We examine the impact of input event sequence density on optical flow learning, as our MDR dataset contains event data with various densities and corresponding dense flow labels. We train SKFlow~\cite{sun2022skflow}, GMA~\cite{jiang2021gma}, FlowFormer~\cite{huang2022flowformer} and KPA-Flow~\cite{luo2022kpa} on the same sequence from our MDR dataset but with different average densities produced by using different threshold $C$, and then test them on MVSEC dataset. 
Figure~\ref{fig:density} shows the results, indicating that these models perform better as the average density of the training set increases. However, their performance diminishes as the average density continues to increase. This phenomenon highlights the importance of selecting an appropriate density for the training set when learning event optical flow.

\paragraph{Ablation for ADM.} 
In order to verify the impact of our proposed ADM module, we conduct ablation experiments by plugging the ADM module into several optical flow network to selectively adjust the densities of the input events. 
We train these networks on both MDR and MVSEC datasets with the same setting except that the ADM module is disabled or not, and test them on MVSEC dataset.
The experiment results are shown in  Table~\ref{tab:comparisonforADM}, where we can notice that our ADM module can bring performance improvement for all supervised methods.


\begin{table}[!t]
	\centering
	\resizebox*{0.99 \linewidth}{!}{
		\begin{tabular}
			{   >{\arraybackslash}p{2.4cm}| 
				>{\centering\arraybackslash}p{0.8cm}| 
				>{\centering\arraybackslash}p{1.2cm} 
				>{\centering\arraybackslash}p{1.2cm}| 
				>{\centering\arraybackslash}p{1.2cm} 
				>{\centering\arraybackslash}p{1.2cm} 
			}
\hline
\multirow{2}{*}{Method } & Train& \multicolumn{2}{c|}{$dt=1$} & \multicolumn{2}{c}{$dt=4$}  \\
&D.Set&EPE & \%Out&EPE & \%Out\\
\hline
\multirow{2}{*}{PWCNet~\cite{Hur:2019:IRR}} & M    & 1.25  & 5.41    & 4.03  & 51.48  \\
          & MDR   & 1.14  & 3.48   & 2.92  & 38.62  \\
\hline
\multirow{2}{*}{E-RAFT~\cite{gehrig2021raft}}  & M     & 1.19  & 4.90     & 3.33  & 39.78  \\
          & MDR   & 0.59  & 0.51   & 2.57  & 30.24  \\
\hline
\multirow{2}{*}{GMFlowNet~\cite{zhao2022GMFlowNet}}  & M     & 1.00  & 3.75    & 3.61  & 42.31  \\
          & MDR   & 0.82  & 1.66   & 2.70  & 31.53  \\
\hline
\multirow{2}{*}{FlowFromer~\cite{huang2022flowformer}} & M     & 0.87  & 3.08    & 3.38  & 41.04  \\
 & MDR   & 0.61  & 0.40    & 2.49  & 28.83  \\
\hline
\multirow{2}{*}{SKFlow~\cite{sun2022skflow}}& M     & 1.07  & 3.97      & 3.41  & 40.87  \\
          & MDR   & 0.59  &\textcolor{blue}{0.33}  & 2.46  & 27.64  \\
\hline
\multirow{2}{*}{GMA~\cite{jiang2021gma}}& M     & 0.88  & 4.05    & 2.99  & 34.80  \\
          & MDR   & \textcolor{blue}{0.58}  & 0.44  & \textcolor{red}{2.19} & \textcolor{red}{23.00} \\
\hline
\multirow{2}{*}{KPAFlow~\cite{luo2022kpa}}& M     & 0.86  & 2.86    & 3.19  & 38.32  \\
          & MDR   & \textcolor{red}{0.58} & \textcolor{red}{0.39} & \textcolor{blue}{2.33}  & \textcolor{blue}{26.10}  \\
\hline
	\end{tabular}}
    \caption{Comparison of training on MVSEC vs. our MDR. Models are evaluated on MVSEC data set for dense optical flow estimation. }
	\label{tab:dataset_comparison}
\end{table}

\paragraph{Ablation for the design of ADM.} 
We conduct ablation experiments to verify the effectiveness of each component in our ADM module, including MDC, MDS, and the two training loss functions $L_{\mathrm{MDC}}$ and $L_{\mathrm{MDS}}$. We train models using the same settings on the MDR dataset and evaluate on the MVSEC dataset to show the individual impact of each component in our ADM module. The results are presented in Table~\ref{tab:ablation}.
Comparison of (a)\&(b) shows that adding only the MDC plugin results in a slight performance gain. Comparison of (b)\&(c) reveals that enabling the density selection function through the MDS module brings a significant improvement. Comparing (c)\&(d) and (d)\&(e), we notice that with the guidance of two loss functions, ADM can learn to selectively choose the best density for optical flow estimation, resulting in a relatively significant improvement.


\begin{table}[t]
	\centering
	\resizebox*{0.99 \linewidth}{!}{
		\begin{tabular}
			{   >{\arraybackslash}p{2.8cm}| 
				>{\centering\arraybackslash}p{0.7cm} 
				>{\centering\arraybackslash}p{0.7cm}| 
				>{\centering\arraybackslash}p{0.7cm} 
				>{\centering\arraybackslash}p{0.7cm}| 
				>{\centering\arraybackslash}p{0.7cm} 
                >{\centering\arraybackslash}p{0.7cm}| 
                >{\centering\arraybackslash}p{0.7cm} 
                >{\centering\arraybackslash}p{0.7cm} 
			}
\hline
\multirow{2}{*}{Method} &\multicolumn{2}{c|}{M($dt=1$)} & \multicolumn{2}{c|}{MDR($dt=1$)}  & \multicolumn{2}{c|}{M($dt=4$)}  & \multicolumn{2}{c}{MDR($dt=4$)} \\
 &EPE & \%Out & EPE & \%Out & EPE & \%Out & EPE & \%Out\\
\hline
    PWCNet~\cite{Hur:2019:IRR} & 1.25  & 5.41  & 1.14  & 3.48  & 4.03  & 51.48  & 2.92  & 38.62  \\
    ADM-PWCNet & 1.07  & 4.52  & 0.76  & 1.48  & 2.95  & 33.31  & 1.94  & 18.74  \\
\hline
    E-RAFT~\cite{gehrig2021raft} & 1.19  & 4.90  & 0.59  & 0.51  & 3.33  & 39.78  & 2.57  & 30.24  \\
    ADM-ERAFT & 0.82  & 3.03  & 0.56  & 0.24  & 2.73  & 30.91  & 1.72  & 13.83  \\
\hline
    GMFlowNet~\cite{zhao2022GMFlowNet} & 1.00  & 3.75  & 0.82  & 1.66  & 3.61  & 42.31  & 2.70  & 31.53  \\
    ADM-GMFlowNet & 0.87  & 3.05  & 0.58  & 0.32  & 2.78  & 31.26  & 1.81  & 14.45  \\
\hline
    FlowFromer~\cite{huang2022flowformer} & 0.87  & 3.08  & 0.61  & 0.40  & 3.38  & 41.04  & 2.49  & 28.83  \\
    ADM-FlowFromer & 0.78  & 2.87  & 0.53  & 0.15  & \textcolor{blue}{2.56}  & \textcolor{blue}{26.57}  & 1.67  & 12.78  \\
\hline
    SKFlow~\cite{sun2022skflow} & 1.07  & 3.97  & 0.59  & 0.33  & 3.41  & 40.87  & 2.46  & 27.64  \\
    ADM-SKFlow & 0.84  & 3.18  & \textcolor{blue}{0.53}  & \textcolor{blue}{0.14}  & 2.67  & 28.17  & 1.69  & 12.61  \\
\hline
    GMA~\cite{jiang2021gma} & 0.88  & 4.05  & 0.58  & 0.44  & 2.99  & 34.80  & 2.19  & 23.00  \\
    ADM-GMA & \textcolor{red}{0.76} & \textcolor{red}{2.65} & 0.54  & 0.22  & \textcolor{red}{2.45} & \textcolor{red}{25.75} & \textcolor{blue}{1.63}  & \textcolor{blue}{11.95}  \\
\hline
    KPAFlow~\cite{luo2022kpa} & 0.86  & 2.86  & 0.58  & 0.39  & 3.19  & 38.32  & 2.33  & 26.10  \\
    ADM-KPAFlow & \textcolor{blue}{0.80}  & \textcolor{blue}{2.81}  & \textcolor{red}{0.51} & \textcolor{red}{0.14} & 2.64  & 28.58  & \textcolor{red}{1.61} & \textcolor{red}{11.83} \\
\hline
	\end{tabular}}
    \caption{Quantitative comparison of whether using ADM. Models are trained on the MVSEC (i.e. M) and MDR training set, and evaluated on the MVSEC test sets for dense optical flow estimation in $dt = 1$ and $dt = 4$ settings.}
	\label{tab:comparisonforADM}
\end{table}

\begin{table}[t]
	\centering
	\resizebox*{0.99 \linewidth}{!}{
		\begin{tabular}
			{   >{\arraybackslash}p{1.0cm}| 
				>{\centering\arraybackslash}p{0.8cm}| 
				>{\centering\arraybackslash}p{0.8cm}| 
				>{\centering\arraybackslash}p{0.8cm}| 
				>{\centering\arraybackslash}p{0.8cm}| 
                >{\centering\arraybackslash}p{0.8cm}| %
                >{\centering\arraybackslash}p{0.7cm} 
                >{\centering\arraybackslash}p{0.7cm}| 
                >{\centering\arraybackslash}p{0.7cm} 
                >{\centering\arraybackslash}p{0.7cm} 
			}
\hline
\multirow{2}{*}{Method} & \multirow{2}{*}{MDC}  & \multirow{2}{*}{MDS} & \multirow{2}{*}{$L_{\mathrm{MDC}}$}  & \multirow{2}{*}{$L_{\mathrm{MDS}}$}  & \multirow{1}{*}{Param.}  & \multicolumn{2}{c|}{$dt=1$}  & \multicolumn{2}{c}{$dt=4$} \\
  & & & & &(M) & EPE & \%Out & EPE & \%Out\\
    \hline
    (a)   & \ding{53} & \ding{53} & \ding{53} & \ding{53} & 6.01  & 0.58  & 0.39  & 2.33  & 26.10 \\
    (b)   & \ding{51} & \ding{53} & \ding{53} & \ding{53} & 7.71  & 0.57  & 0.33  & 2.20   & 23.78 \\
    (c)   & \ding{51} & \ding{51} & \ding{53} & \ding{53} & 7.72  & 0.54  & 0.26  & 1.92  & 18.26 \\
    (d)   & \ding{51} & \ding{51} & \ding{51} & \ding{53} & 7.72  & \textcolor{blue}{0.52}  & \textcolor{blue}{0.16}  & \textcolor{blue}{1.66}  & \textcolor{blue}{13.29}  \\
    (e)   & \ding{51} & \ding{51} & \ding{51} & \ding{51} & 7.72  & \textcolor{red}{0.51 } & \textcolor{red}{0.14 } & \textcolor{red}{1.61} & \textcolor{red}{11.83} \\
\hline
	\end{tabular}}
    \caption{Ablation study. Models are trained on the MDR training set, and evaluated on the MVSEC test sets for dense optical flow estimation in $dt = 1$ and $dt = 4$ settings.}
	\label{tab:ablation}
\end{table}


\section{Conclusion}
In this work, we have created a rendered dataset for event-flow learning. Indoor and outdoor virtual scenes have been created using Blender with rich scene contents. Various camera motions are placed for the capturing of the virtual world, which can produce frames as well as accurate flow labels. The event values are generated by render high frame rate videos between two frames. In this way, the flow labels and event values are physically correct and accurate. The rendered dataset can adjust density of events by modifying the event trigger threshold. We have introduced a novel adaptive density module (ADM), which has shown its effectiveness by plugin into various event-flow pipelines. When trained on our dataset, previous approaches can improve their performances constantly.  

{\small
\bibliographystyle{ieee_fullname}
\bibliography{egbib}
}

\end{document}